\documentclass[letterpaper]{article}
\usepackage{aaai2027}
\usepackage[hyphens]{url}
\usepackage{graphicx}
\usepackage[numbers,sort&compress]{natbib}
\usepackage{caption}
\usepackage{booktabs}
\usepackage{amsmath}
\usepackage{amssymb}
\title{PIA: Agentic Physics-Informed Runtime Anomaly Detection}
\author{
    Ali Mohammadi Ruzbahani\textsuperscript{\rm 1},
    Abbas Yazdinejad\textsuperscript{\rm 2},
    Hadis Karimipour\textsuperscript{\rm 1}
}
\affiliations{
    \textsuperscript{\rm 1}Smart Cyber-Physical Systems (SCPS) Lab, University of Calgary, Calgary, Alberta, Canada\\
    \textsuperscript{\rm 2}Decentralized Cybersecurity and Artificial Intelligence(DCAILab), University of Regina, Regina, Saskatchewan, Canada\\
    ali.mohammadiruzbaha@ucalgary.ca, Abbas.Yazdinejad@uregina.ca, hadis.karimipour@ucalgary.ca
}

\begin{document}
\maketitle
\begin{abstract}
Runtime anomaly detectors for cyber-physical systems are either learned as black boxes from normal data, which miss stealthy manipulations that keep every signal inside its normal range, or derived by hand from physical models, which requires expert effort for every plant. We introduce physics-informed agents (PIA), in which a large language model agent writes the detector itself as a set of physical residuals, such as pump characteristics, tank mass balances, and controller logic, and refines every candidate against normal operating data through an evaluation workbench before deployment. The result is a compact, human-readable detector that runs in milliseconds and names the component behind each alarm. On the BATADAL water distribution attack benchmark, PIA with Claude Opus 5.5 reaches an F1 of 0.878 on the test attacks against 0.504 for the strongest of six statistical and deep baselines, keeps detecting attacks concealed by replayed and offset sensor readings, and attributes every detected test attack to the attacked component. Ablations show that testing hypotheses against data is decisive: agents that write residuals from the plant description alone lose up to 0.28 F1, whereas agents that see only anonymized signals recover most of the physical structure.
\end{abstract}
\section{Introduction}

Detectors that watch cyber-physical systems at run time must flag attacks that keep every individual signal inside its normal range. Detectors learned from normal data, such as autoencoders and recurrent forecasters, model the joint distribution of all sensors without knowing which relations are physically binding. A replayed tank level or an offset pressure reading that remains statistically plausible therefore passes unnoticed, and an alarm, when raised, gives no indication of where the problem lies. Physics-based residual detectors avoid both issues by checking relations that must hold in any normal operation \citep{giraldo2018survey}, but every residual has to be derived, parameterized, and tuned by an engineer for the specific plant. PIA removes this bottleneck by delegating the engineering work to a language model agent that already holds broad knowledge of hydraulics, machines, and control logic, and that tests every relation it proposes against normal operating data before the relation enters the detector. The agent thus produces the kind of detector a domain expert would write, with the sensitivity and interpretability of physical residuals, after a few minutes of automated design.

Recent advances have sought to bridge data-driven monitoring
and physical reasoning in cyber-physical systems. Wu et
al.~\citep{wu2023physics} incorporated physical constraints
into graph-based temporal learning for industrial anomaly
detection, while Ashraf et al.~\citep{ashraf2024physics}
demonstrated how hydraulic principles can guide graph neural
networks in modeling water distribution systems. Beyond
these applications, Shojaee et al.~\citep{shojaee2025llmsr}
showed that large language models can propose interpretable
scientific equations whose parameters are subsequently
optimized against observations. Together, these developments
motivate a shift from learning correlations or manually
specifying physical constraints toward automatically
constructing physically meaningful models. However,
translating this capability into lightweight, interpretable
runtime attack detectors remains a distinct challenge. PIA
addresses this challenge by using an LLM agent to propose,
evaluate, and refine physical residuals directly from plant
descriptions and normal operating data.

The agent receives a short natural language description of the plant and a workbench over normal data. It proposes residuals in a small expression language, observes how each candidate fits, how often it fires on normal data it was not fitted on, and how strongly it reacts to synthetic sensor manipulations, and revises the set until the combined detector is quiet on normal data and covers every sensor and actuator. The submitted residuals are then fitted without further involvement of the model, so the deployed detector is deterministic, executes in milliseconds, and localizes each alarm through the residual that carries it.

Our contributions are as follows.
\begin{itemize}
\item We formulate runtime anomaly detection for cyber-physical systems as agentic program synthesis, in which a language model writes a detector of physical residuals and grounds it in normal data through an evaluation workbench.
\item We design the residual language and the workbench feedback that turn agent hypotheses into a deterministic, interpretable detector with component-level localization.
\item We evaluate PIA with two frontier models on the BATADAL water distribution benchmark against six established detectors, and release all prompts, transcripts, residual sets, checkpoints, and scores.
\item We show through ablations that empirical grounding, rather than prior knowledge alone, drives detection quality, and that agents rediscover physical structure from anonymized signals.
\end{itemize}
\section{Related Work}

Physics-based attack detection compares sensor readings with the predictions of a system model and raises an alarm when the residual exceeds a bound; \citet{giraldo2018survey} survey these detectors and the stateless and stateful residual tests they rely on. Their strength is that a stealthy attacker must keep many coupled readings consistent at once, while their cost is a model built by hand for every plant. Statistical process monitoring with principal component analysis \citep{qin2003spm}, isolation forests \citep{liu2008isolation}, recurrent forecasters \citep{hundman2018lstm}, autoencoders for water networks \citep{taormina2018deep}, and adversarially trained autoencoders \citep{audibert2020usad} learn normal behaviour without physical structure and serve as our baselines. The BATADAL competition \citep{taormina2018battle} established a common benchmark for attacks on water distribution networks.

Physics-informed machine learning embeds known laws into learning \citep{karniadakis2021piml}, and sparse regression discovers governing equations from data \citep{brunton2016sindy}; both assume that the candidate terms are specified by the modeller. Language models paired with automatic evaluators have discovered new programs through iterative search \citep{romeraparedes2024funsearch}. PIA connects these lines: the language model supplies the physical hypotheses and the terms of each relation, and the workbench supplies the empirical feedback that an engineer or a symbolic regression would otherwise provide.

Recent work has also investigated the security and adaptability of
learning-based systems under adversarial conditions. Mohammadi
Ruzbahani et al.~\cite{MOHAMMADIRUZBAHANI2026104495}
studied quantum-adaptive ensemble attacks against robust deep
neural networks, while hierarchical quantum reinforcement learning
has been explored for defending federated learning against
multi-vector poisoning attacks~\cite{Ruzbahani2026HQRL}.
Complementary directions include decentralized collaborative
learning without model sharing under adversarial heterogeneity
\cite{ruzbahani2026trustless} and approximate machine unlearning
using Kronecker-factored tensor networks
\cite{ruzbahani2026tensorunlearn}. These studies address
adversarial robustness, distributed learning, and model
maintenance, whereas PIA focuses on synthesizing interpretable
physical consistency checks for runtime attack detection.

Large language models are increasingly being investigated
as components of adaptive learning and security workflows.
FedReason~\cite{ruzbahani2026fedreasonlargelanguagemodels}
explores LLMs as weight-space meta-aggregators in heterogeneous
federated learning. In agentic environments, memory poisoning
introduces another attack surface, as investigated by
Yazdinejad and Karimipour~\cite{a1}, while federated
unlearning in agentic energy systems raises questions about
security and governance~\cite{a2}. Unlike methods that
operate on model parameters, persistent memories, or
distributed training processes, PIA employs an LLM during
detector construction to propose and empirically test
physical residuals. The resulting detector is fixed,
interpretable, and independent of the LLM during runtime.

\section{Problem Formulation}

A cyber-physical system is observed through a multivariate stream $x_t \in \mathbb{R}^d$ that combines continuous sensor readings, such as levels, flows, and pressures, with binary actuator states reported by the controllers. Each signal belongs to one or more physical components in a set $\mathcal{K}$ of tanks, pumps, valves, and junctions. Under normal operation the stream obeys physical laws and control rules that couple signals across components. An attack manipulates the process, the control logic, or the readings delivered to the supervisory system, and a capable attacker conceals its effect by replaying or offsetting a subset of readings so that each manipulated signal remains inside its normal range.

The defender has a segment of normal operation $\mathcal{N} = (x_1, \dots, x_n)$ that contains no attacks and a natural language description $P$ of the plant as found in its documentation, but no examples of attacks. The task is to construct, before deployment, a detector that maps the history $x_{\le t}$ at every time step to an alarm $a_t \in \{0, 1\}$ and, for every alarm episode, to a component $c \in \mathcal{K}$ responsible for it. The detector must be causal, so that it can run online, and cheap enough to evaluate at the sampling rate of the supervisory system.

We cast this task as program synthesis. A detector is a program $\phi$ in a restricted language of residual expressions, together with parameters $\theta$ fitted on normal data. The design objective is
\begin{equation}
\phi^{\star} = \arg\max_{\phi \in \Phi} \; \mathrm{Sens}(\phi, \theta_{\phi}) \quad \text{s.t.} \quad \mathrm{FAR}(\phi, \theta_{\phi}) \le \alpha,
\end{equation}
where $\theta_{\phi}$ is fitted on $\mathcal{N}$, $\mathrm{FAR}$ is the fraction of normal time steps that raise an alarm, and $\mathrm{Sens}$ is the sensitivity of the detector to manipulations of the plant. Neither the attacks nor $\mathrm{Sens}$ are available during design, so the designer must reason about which relations a manipulation would violate. This is the step that physical knowledge supplies and that purely statistical learning lacks, and it is the step we delegate to a language model agent equipped with $P$ and with tools that estimate the false alarm rate and a proxy of sensitivity on normal data.
\begin{figure*}[t]
\centering
\includegraphics[width=\textwidth]{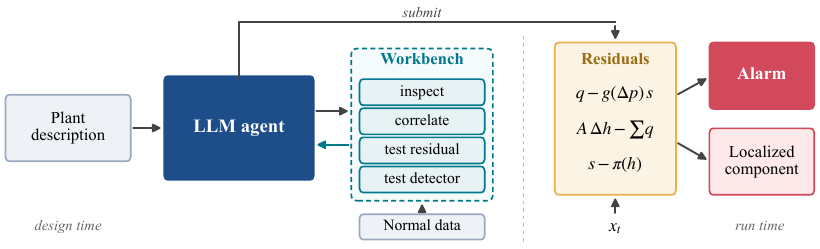}
\caption{Overview of PIA.}
\label{fig:overview}
\end{figure*}

\section{Method}

PIA separates a design phase, in which the agent writes residuals, from a run-time phase, in which the residuals are evaluated without the model (Figure~\ref{fig:overview}). A residual $k$ consists of a target expression $g_k$ and an optional list of input expressions $h_k$ over the signals. Expressions combine signals with arithmetic, comparisons, logical operators, lags, differences, rolling means and sums, and conditional selection, which suffices to state flow and pressure relations, discretized mass balances, and the hysteresis logic of programmable controllers. The raw residual is
\begin{equation}
r_{k,t} = g_k(x_{\le t}) - \theta_k^{\top} \begin{bmatrix} h_k(x_{\le t}) \\ 1 \end{bmatrix},
\end{equation}
where $\theta_k$ is fitted by least squares on normal data. The regression absorbs constants the agent cannot know, such as tank cross sections, pump gains, or friction coefficients, while the structure of the relation comes from the agent. Each residual is standardized by its median and median absolute deviation on normal data, its absolute value is averaged over a short window $w_k$ chosen by the agent, and the result is divided by its upper quantile on normal data, giving $u_{k,t}$. All residuals thus share a common scale on which $1$ marks the edge of normal behaviour. The detector score and alarm are
\begin{equation}
s_t = \max_k u_{k,t}, \qquad a_t = \mathbf{1}\left[s_t > \tau\right],
\end{equation}
with $\tau$ an upper quantile of $s_t$ on normal data. Each residual lists the physical components it involves, and an alarm episode is attributed to the components of the residual with the largest accumulated $u_{k,t}$ over the episode.

The agent designs the set $\{(g_k, h_k, w_k)\}$ in a multi-turn session. Its system prompt explains the residual form, the expression language, and the principle that residuals should be nearly constant in normal operation, followed by the plant description $P$, which gives the meaning and units of each signal, the topology of pumps, tanks, and valves, tank dimensions, and the switching rules of the controllers. The agent never sees anomalous data. It interacts with a workbench of four tools. The first summarizes every signal on normal data, and the second correlates arbitrary expressions with all signals. The third fits a candidate residual on one part of the normal data and evaluates it on another, returning its explained variance, its false alarm rate, the ratio between evaluation and fitting tail quantiles, which exposes drifting or overfitted relations, and the fraction of a synthetic manipulation it detects when each involved signal is offset by half a standard deviation or, for binary actuators, inverted. The fourth assembles a candidate set into the full detector and reports its false alarm rate, the residuals responsible for false alarms, and the signals that no residual covers. The session ends when the agent submits between 4 and 24 residuals.

This division of labour exploits complementary strengths. The language model contributes the hypothesis space, proposing relations that a generic learner would not consider, such as a pump flow that must vanish when its status is off, or a level change that must equal inflow minus withdrawals. The workbench contributes falsification, rejecting relations that hold in textbooks but not in the data and exposing residuals whose noise would raise the common threshold. Because the deployed detector is a fixed set of fitted expressions, its behaviour can be audited line by line, and its run-time cost is that of evaluating a few dozen arithmetic expressions per time step.
\section{Experimental Setup}

BATADAL \citep{taormina2018battle} contains hourly SCADA data of the C-Town water distribution network with 43 signals: seven tank levels, flows and statuses of eleven pumps and one valve, and twelve junction pressures at pump and valve stations. The competition provides a one-year normal dataset, a six-month Dataset 2 with seven attacks, and a three-month test dataset with seven further attacks. All detectors and all agents use only the normal dataset; Dataset 2 and the test dataset are used exclusively for evaluation, with labels taken from the attack schedules published by the organizers. The attacks alter level readings sent to controllers, change control thresholds, activate pumps, or reduce pump speed, and most conceal their effect by replaying or offsetting SCADA readings. Their duration ranges from 24 to 110 hours. We report point-wise precision (P), recall (R), F1, average precision of the score (AP), and the competition score $S$, which averages the normalized time to detection with the mean of true positive and true negative rates \citep{taormina2018battle}.

The baselines are PCA monitoring with combined $T^2$ and $Q$ statistics \citep{qin2003spm}, isolation forest (IForest) \citep{liu2008isolation}, a dense autoencoder (AE) \citep{taormina2018deep}, an LSTM forecaster \citep{hundman2018lstm}, an LSTM autoencoder (LSTM-AE), and USAD \citep{audibert2020usad}. All methods see the same normal data and share the score smoothing and the quantile rule for the threshold, so that differences reflect the scores themselves. Stochastic baselines are trained with five seeds. PIA uses Claude Sonnet 5.5 and Claude Opus 5.5 with five independent sessions per model and configuration, each limited to 40 turns.
\begin{table*}[t]
\centering
\small
\setlength{\tabcolsep}{4.2pt}
\begin{tabular}{@{}lccccc@{\hspace{10pt}}ccccc@{}}
\toprule
 & \multicolumn{5}{c}{Dataset 2} & \multicolumn{5}{c}{Test} \\
\cmidrule(lr){2-6}\cmidrule(l){7-11}
Method & P & R & F1 & AP & $S$ & P & R & F1 & AP & $S$ \\
\midrule
PCA & \textbf{.901} & .482 & .628 & .697 & .769 & \textbf{.904} & .349 & .504 & .713 & .784 \\
IForest & .541{\scriptsize$\pm$.02} & .069{\scriptsize$\pm$.01} & .122{\scriptsize$\pm$.01} & .261{\scriptsize$\pm$.01} & .490{\scriptsize$\pm$.05} & .590{\scriptsize$\pm$.08} & .026{\scriptsize$\pm$.00} & .049{\scriptsize$\pm$.01} & .269{\scriptsize$\pm$.00} & .445{\scriptsize$\pm$.02} \\
AE & .874{\scriptsize$\pm$.02} & .472{\scriptsize$\pm$.01} & .612{\scriptsize$\pm$.01} & .688{\scriptsize$\pm$.02} & .808{\scriptsize$\pm$.01} & .880{\scriptsize$\pm$.04} & .329{\scriptsize$\pm$.03} & .478{\scriptsize$\pm$.03} & .664{\scriptsize$\pm$.01} & .776{\scriptsize$\pm$.02} \\
LSTM & .889{\scriptsize$\pm$.01} & .390{\scriptsize$\pm$.07} & .539{\scriptsize$\pm$.06} & .723{\scriptsize$\pm$.01} & .715{\scriptsize$\pm$.08} & .885{\scriptsize$\pm$.04} & .191{\scriptsize$\pm$.04} & .311{\scriptsize$\pm$.05} & .682{\scriptsize$\pm$.02} & .557{\scriptsize$\pm$.10} \\
LSTM-AE & .704{\scriptsize$\pm$.02} & .459{\scriptsize$\pm$.02} & .555{\scriptsize$\pm$.02} & .603{\scriptsize$\pm$.02} & .754{\scriptsize$\pm$.05} & .766{\scriptsize$\pm$.03} & .329{\scriptsize$\pm$.03} & .460{\scriptsize$\pm$.03} & .588{\scriptsize$\pm$.01} & .698{\scriptsize$\pm$.02} \\
USAD & .531{\scriptsize$\pm$.10} & .086{\scriptsize$\pm$.04} & .147{\scriptsize$\pm$.06} & .230{\scriptsize$\pm$.06} & .435{\scriptsize$\pm$.06} & .682{\scriptsize$\pm$.02} & .171{\scriptsize$\pm$.02} & .273{\scriptsize$\pm$.03} & .406{\scriptsize$\pm$.02} & .540{\scriptsize$\pm$.05} \\
\midrule
PIA (Sonnet 5.5) & .834{\scriptsize$\pm$.03} & .713{\scriptsize$\pm$.03} & .769{\scriptsize$\pm$.02} & .775{\scriptsize$\pm$.07} & .911{\scriptsize$\pm$.01} & .862{\scriptsize$\pm$.05} & .640{\scriptsize$\pm$.07} & .731{\scriptsize$\pm$.03} & .729{\scriptsize$\pm$.03} & .876{\scriptsize$\pm$.01} \\
PIA (Opus 5.5) & .838{\scriptsize$\pm$.04} & \textbf{.812}{\scriptsize$\pm$.04} & \textbf{.824}{\scriptsize$\pm$.02} & \textbf{.854}{\scriptsize$\pm$.02} & \textbf{.931}{\scriptsize$\pm$.01} & .885{\scriptsize$\pm$.07} & \textbf{.874}{\scriptsize$\pm$.04} & \textbf{.878}{\scriptsize$\pm$.03} & \textbf{.899}{\scriptsize$\pm$.02} & \textbf{.927}{\scriptsize$\pm$.03} \\
\bottomrule
\end{tabular}
\caption{BATADAL results, mean and standard deviation over five runs.}
\label{tab:main}
\end{table*}

\section{Results}

Table~\ref{tab:main} compares PIA with the baselines on both attack datasets. On the test attacks, PIA with Opus raises F1 from 0.504 for the best baseline to 0.878 and AP from 0.713 to 0.899 while keeping precision on par with PCA, and the competition score rises from 0.784 to 0.927. Dataset 2 shows the same ordering, with F1 rising from 0.628 to 0.824. The gain comes from recall, which grows from 0.349 to 0.874 on the test attacks: the baselines catch the onset of most attacks but lose them once concealment takes effect, whereas the residuals keep firing for the full attack duration. Opus sessions detected all seven test attacks in four of five cases and six in the fifth, and in every session the residual carrying each detected attack involved the attacked component. Sonnet reaches 0.731 and 0.769 F1, which still exceeds every baseline by a wide margin.

\begin{figure*}[t]
\centering
\includegraphics[width=\textwidth]{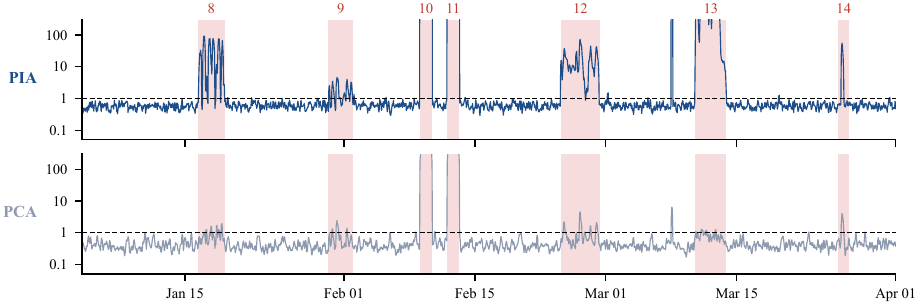}
\caption{Normalized scores on the BATADAL test period; shaded spans are attacks.}
\label{fig:timeline}
\end{figure*}

\begin{figure}[t]
\centering
\includegraphics[width=\columnwidth]{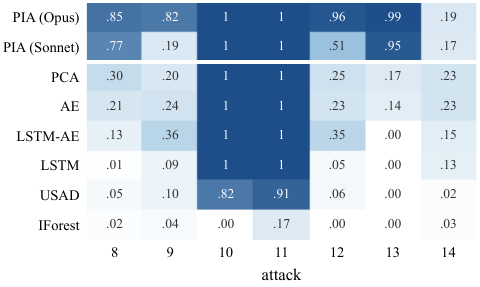}
\caption{Fraction of each test attack under alarm.}
\label{fig:attacks}
\end{figure}

The residual sets written by Opus share a recognizable structure. Every session wrote pump characteristics that relate the flow of each pump to its status and to the pressure rise between its suction and discharge junctions, a residual stating that pumps kept in reserve carry neither flow nor status, tank balances or tank head relations driven by the flows of the feeding pumps and valve, and head relations between stations that share a suction main or draw from the same tank. Four of the five sessions also encoded the controller switching rules. Figures~\ref{fig:timeline} and~\ref{fig:attacks} show how this structure translates into detection. Attacks 10 and 11 switch on a reserve pump without concealment and are flagged throughout by nearly every detector. Attacks 8, 9, 12, and 13 conceal their effect, and here no baseline keeps the alarm on for more than 36\% of any attack, while PIA with Opus covers between 82\% and 99\% of each. In attack 13 the alarm is carried by the relation between the suction heads of two neighbouring stations, since the replayed readings of the targeted station no longer agree with the live readings next to it. In attacks 9 and 12 it is carried by relations around tank T2 that tie its level to the heads at the valve and the pump station connected to it, which the manipulated level violates. Attack 14 alters the T4 level on its way to a controller while the reported T4 level stays in its normal band, and all detectors, PIA included, flag it only around a single transient.

\begin{figure}[t]
\centering
\includegraphics[width=\columnwidth]{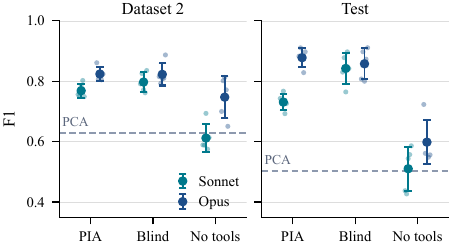}
\caption{Ablation of the plant description and the workbench; dashed lines mark the best baseline.}
\label{fig:ablation}
\end{figure}

Figure~\ref{fig:ablation} isolates the two sources of information available to the agent. In the blind configuration the plant description is reduced to one sentence and all signals are renamed to anonymous identifiers, so the agent must infer the system from the workbench alone. In the configuration without tools the agent keeps the full description but submits residuals without seeing any data. Removing the workbench costs 0.28 F1 for Opus and 0.22 for Sonnet on the test attacks and lowers F1 on Dataset 2 as well. Blind agents, in contrast, reach 0.858 and 0.842 F1 on the test attacks. Their transcripts show the agent recognizing pumps from pairs of binary and continuous signals, matching pumps to junctions through correlations, and then writing head characteristics and tank balances for the anonymous signals. For Opus, the description raises localization accuracy on the test attacks from 0.714 to 1.0, because named components let the agent tie every residual to the equipment it monitors.

Designing a detector is a one-time cost. An Opus session issued 52 tool calls on average, consumed 364 thousand input and 23 thousand output tokens, and finished in 194 seconds; a Sonnet session needed 38 calls and 91 seconds. The deployed detector holds about twenty fitted expressions, is fitted in 0.13 seconds, and scores the three-month test period in 33 milliseconds on a CPU, without a GPU and without calls to the model.
\section{Discussion}

The results point to a specific mechanism behind the advantage of PIA. Learned detectors summarize normal behaviour by its dominant modes of variation, and the strongest variation in a water network comes from daily demand and from the switching of pumps. A concealed attack that replays a plausible level trace or offsets a reading by a moderate amount stays within these modes and produces only a small reconstruction or forecast error. A physical residual instead isolates a relation whose normal variation is small, such as the head rise of a pump at a given flow or the agreement between two pressures on one suction main. The same manipulation that is negligible against the total variance of the plant is large against the variance of such a relation, which explains why the residuals reach scores one to two orders of magnitude above the threshold on attacks that the baselines barely register. To stay hidden, the attacker would have to forge every reading that participates in a violated relation, and because the agent covers each component with several overlapping relations, a consistent forgery has to extend well beyond the attacked component.

The separation between agentic design and deterministic
execution also suggests directions for maintaining trustworthy
detectors over time. Prior research on adversarial learning
\cite{MOHAMMADIRUZBAHANI2026104495,Ruzbahani2026HQRL},
decentralized collaboration~\cite{ruzbahani2026trustless},
and machine unlearning~\cite{ruzbahani2026tensorunlearn,a2}
highlights related challenges in protecting and updating
learning-based systems. For PIA, an important future question
is how residual programs should be revised when plant
configuration, operating conditions, or sensor reliability
change. Such revision would require validating newly proposed
relations against normal operating data while preserving
the interpretability and low runtime cost of the detector.
These extensions remain outside the present evaluation.

The ablations clarify the role of the language model. Its contribution is not a closed-form model of the plant; the agent working from the description alone writes sensible relations that are too loose for this network, and the agent working from anonymous signals recovers nearly the same detection quality. The decisive contribution is the search over relational structure: proposing which signals should be coupled and in what functional form, and revising the proposal when the workbench shows that a relation is noisy or drifts. This is the step that equation discovery leaves to the modeller and that statistical learners leave implicit. The plant description adds a semantic layer on top of this search, so that each residual is attached to named equipment and alarms point to the component under attack.

These properties fit the workflow of plant operators. A detector expressed as a short list of physical relations can be reviewed before deployment by an engineer who does not work with machine learning, a false alarm can be traced to the relation that fired, and a relation that no longer holds after a change in the plant can be replaced without retraining the rest. Because the language model participates only in design, its cost is paid once per plant and its outputs are fixed and auditable at run time. Natural extensions are revision of the residual set when the plant changes, design sessions in which the agent consults engineering documentation or network models directly, and attribution that goes beyond the component to the type of manipulation.
\section{Conclusion}

PIA turns a language model into the designer of a runtime detector for cyber-physical systems. By writing physical residuals and testing each one against normal operation, the agent produces detectors that expose concealed attacks missed by statistical and deep baselines, localize them to the affected component, and run at negligible cost. Agentic program synthesis thus offers a practical route to interpretable monitoring of plants for which hand-built physical models were previously the only option.

\section*{AI-use disclosure}
During the preparation of this work, the authors used AI in order to improve the readability of the code and to debug it, to improve the text, to correct grammatical errors, to trim the text, and to improve the visualization of the figures.
The initial hypothesis and model originated with the authors,
who are responsible for the final manuscript, references
and results.
\bibliography{refs}
\end{document}